\pdfoutput=1

\documentclass[11pt]{article}

\usepackage[]{acl}

\usepackage{times}
\usepackage{latexsym}

\usepackage[T1]{fontenc}

\usepackage[utf8]{inputenc}
\usepackage{todonotes}

\usepackage{microtype}
\usepackage{amsmath,amssymb}
\usepackage{epsfig,graphicx,subfigure,caption}
\usepackage{algpseudocode}
\usepackage[normalem]{ulem}
\usepackage{amsmath}
\usepackage[linesnumbered,algoruled,boxed,noend]{algorithm2e}
\usepackage{xcolor}
\usepackage{color, colortbl}
\usepackage{multirow,booktabs, hhline}
\usepackage{amsmath, bm}
\usepackage{wrapfig}
\usepackage{listings}
\usepackage{adjustbox}
\usepackage{makecell}

\usepackage{url}
\usepackage[T1]{fontenc}
\usepackage{tikz}
\usepackage{soul}
\usepackage{enumitem}
\usepackage[many]{tcolorbox}
\usepackage{pifont}
\newcommand{\cmark}{\ding{51}}
\newcommand{\xmark}{\ding{55}}

\setlist[enumerate]{itemsep=0pt, parsep=0pt}
\setlist[itemize]{itemsep=4pt, parsep=0pt}

\definecolor{darkgreen}{RGB}{0,100,0}
\definecolor{darkred}{RGB}{153,0,0}
\definecolor{darkorange}{RGB}{184,85,0}
\definecolor{lightyellow}{RGB}{255,255,224}
\definecolor{loss}{RGB}{196,118,108}
\definecolor{win}{RGB}{164,193,136}

\definecolor{blush}{rgb}{0.87, 0.36, 0.51}

\newcommand{\ours}{\textsc{TICL}\xspace}

\title{Tuning-Free Personalized Alignment via Trial-Error-Explain \\ 
In-Context Learning}

\author{
    Hyundong Cho\textsuperscript{1}\Thanks{Work was done while HC was an intern at Amazon AGI.},~
    Karishma Sharma\textsuperscript{2},~
    Nicolaas Jedema\textsuperscript{2},~
    Leonardo F. R. Ribeiro\textsuperscript{2},~ 
    \\
    \textbf{
    Alessandro Moschitti\textsuperscript{2},~
    Ravi Krishnan\textsuperscript{2},~
    Jonathan May\textsuperscript{1}
    }
\\
\textsuperscript{1}University of Southern California, Information Sciences Institute~
\textsuperscript{2}Amazon AGI
\\
{\small \texttt{hd.justincho@gmail.com}}
}

\begin{document}
\maketitle

\begin{abstract}

Language models are aligned to the collective voice of many, resulting in generic outputs that do not align with specific users' styles. 
In this work, 
we present \textit{Trial-Error-Explain In-Context Learning} (\textbf{\ours}), a tuning-free method that personalizes language models for text generation tasks with fewer than 10 examples per user. 
\ours iteratively expands an in-context learning prompt via a \textit{trial-error-explain} process, adding model-generated negative samples and explanations that provide fine-grained guidance towards a specific user's style.
\ours achieves favorable win rates on pairwise comparisons with LLM-as-a-judge up to 91.5\% against the previous state-of-the-art and outperforms competitive tuning-free baselines for personalized alignment tasks of writing emails, essays and news articles. 
Both lexical and qualitative analyses show that the negative samples and explanations enable language models to learn stylistic context more effectively and overcome the bias towards structural and formal phrases observed in their zero-shot outputs.  
By front-loading inference compute to create a user-specific in-context learning prompt that does not require extra generation steps at test time, \ours presents a novel yet simple approach for personalized alignment.

\end{abstract}

\section{Introduction}

\begin{figure}[ht]
    \centering
    \includegraphics[width=\linewidth]{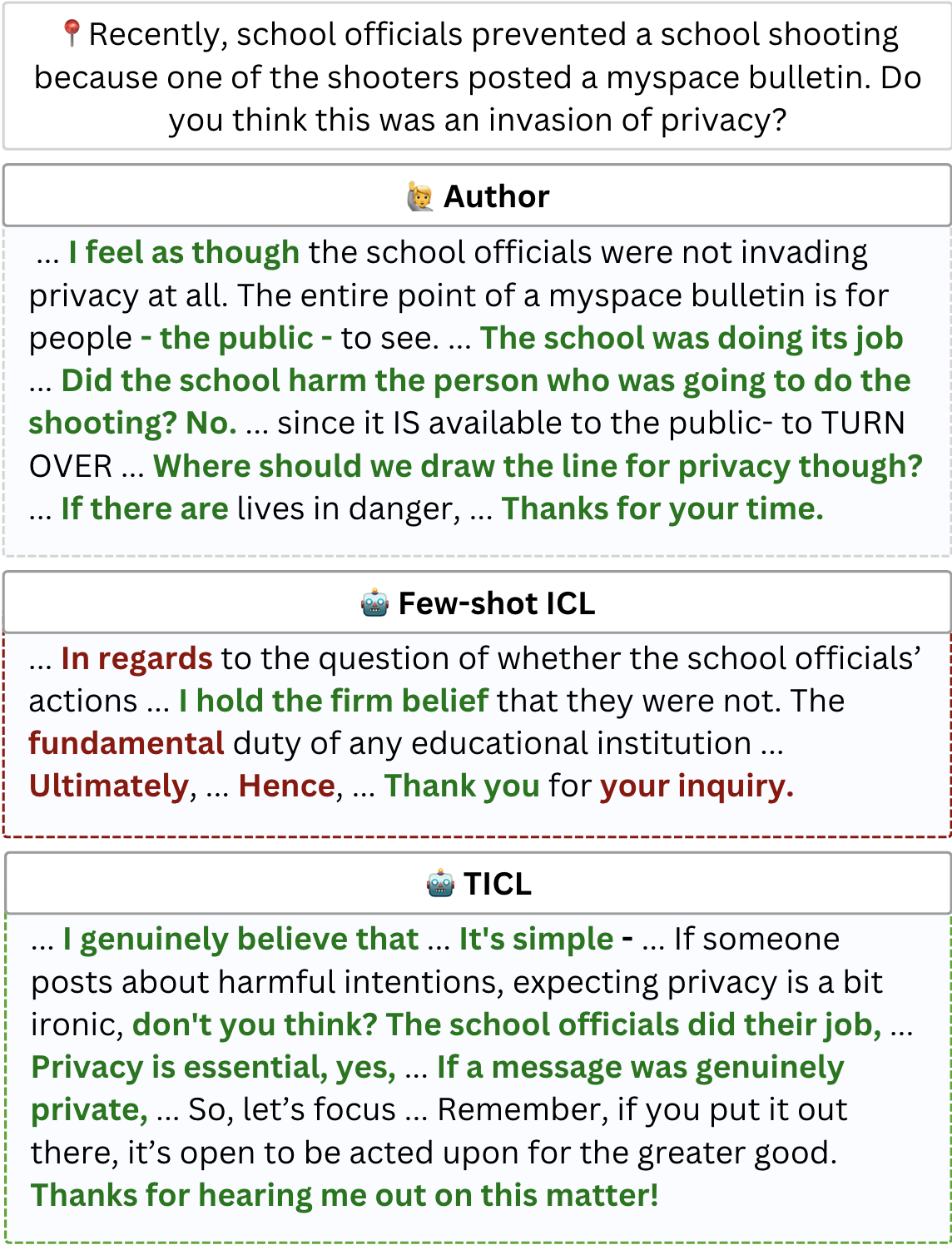}
    \caption{\textit{Top:} The author's text contains rhetorical questions and more colloquial and informal phrases.
    \textit{Middle}: Few-shot in-context learning is insufficient for adapting to an author's style and model biases for \textcolor{darkred}{formal phrases persist.}
    \textit{Bottom:} \ours helps overcome model biases and applies the \textcolor{darkgreen}{author's style} more consistently. 
    Model outputs are with \texttt{gpt-4o-2024-0806}
    }
    \label{fig:motivating_example} 
\end{figure}

Large language models (LLMs) are biased towards generic outputs as they are trained to align to an aggregate preference to be generally useful~\cite{padmakumar2024does,rafailov2024direct, lee2023rlaif, wang-etal-2023-self-instruct}. 
Yet end users often have specific needs that call for personalized text generation, such as writing an email~\cite{kumar2024longlampbenchmarkpersonalizedlongform, salemi2023lamp}. 
Hence, adapting LLMs for personalized text generation has drawn a growing interest~\cite{ jang2023personalized, shaikh2024show, mysore2023pearl, li2024learning}, but prior work on personalized text generation either rely on large amounts of personal data~\cite{li2024learning, li2023teach, mysore2023pearl} or parameter updates that result in cumbersome per-user\footnote{We use the term ``user'' and ``author'' interchangeably throughout this work. A user is anyone who provides personally written text that style can be learned from.} model parameters~\cite{shaikh2024show, liu-etal-2023-recap}.

These are critical limitations. 
Personalization is impractical when premised on users providing ample data because most are not willing to exert such effort~\cite{ tetard2009lazy} or are concerned about privacy~\cite{plant2022you}.
In addition, fine-tuning may not even be a viable option. The best performing LLMs are most often only accessible via APIs or too large that fine-tuning them is prohibitive even if they are open-sourced~\cite{touvron2023llama, firsich-rios-2024-gpt4, TheC3}.
Can we develop a method that overcomes these limitations such that it is effective with small amounts of data and requires no parameter updates? 

To address this problem, we propose \textbf{T}rial-Error-Explain \textbf{I}n-\textbf{C}ontext \textbf{L}earning (\textbf{\ours}), a \textit{tuning-free} adaptation of Trial-and-Error Fine-tuning (TEFT)~\cite{shaikh2024show,gulcehre2023reinforced,song-etal-2024-trial}.
The key idea of TEFT is to iteratively generate synthetic negative samples to create a preference dataset and apply a preference optimization method such as DPO~\cite{rafailov2024direct} to learn a more nuanced preference even with a small amount of initial training data. 
\ours takes a similar approach but only relies on scaling inference compute to expand an in-context learning (ICL) prompt to include prior model-generated negative samples (\textit{trial-error}) and corresponding explanations that describe their shortcomings (\textit{explain}) to better understand a user's personal style. 

We evaluate the effectiveness of \ours with GPT-4o and Claude 3 Sonnet on two personalized writing datasets consisting of emails, essays, and news articles.  
Using an extensively verified LLM-as-a-judge setup as the evaluator for comparing style similarity between two pairs of generated text with a user's text, we find that \ours achieves higher win rates (i.e. more frequently considered more similar in pairwise comparisons) than all other competitive prompting-based baselines, including Chain-of-Thought~\cite{wei2022chain} and OPRO~\cite{yang2024large} for models. 
\ours also achieves an average win rate of 76.6\% against the previous state-of-the-art~\cite{shaikh2024show}.
We ablate \ours extensively and find that, while every procedure in \ours contributes to stronger performance, explanations are crucial, as it is responsible for up to 77\% of the gains relative to a simple ICL baseline.

Lastly, we also share lexical analysis that shed light on the performance gaps between various models and methods. 
As illustrated in \autoref{fig:motivating_example}, compared to ICL,
we find \ours more reliably reduces LLM bias towards structural and formal phrases, e.g.,  ``\textit{additionally}'' and ``\textit{therefore}'', and instead use more colloquial ones, e.g., ``\textit{so why}'' and ``\textit{honestly.}'' 
We also discover that Claude 3 Sonnet's consistent dominance over GPT-4o is attributed to Claude 3 Sonnet more reliably generating phrases that elicit opinions, such as ``\textit{believe}'' and ``\textit{feel that},'' which are frequently seen in the users' texts.

\begin{figure*}
    \centering
    \includegraphics[width=\linewidth]{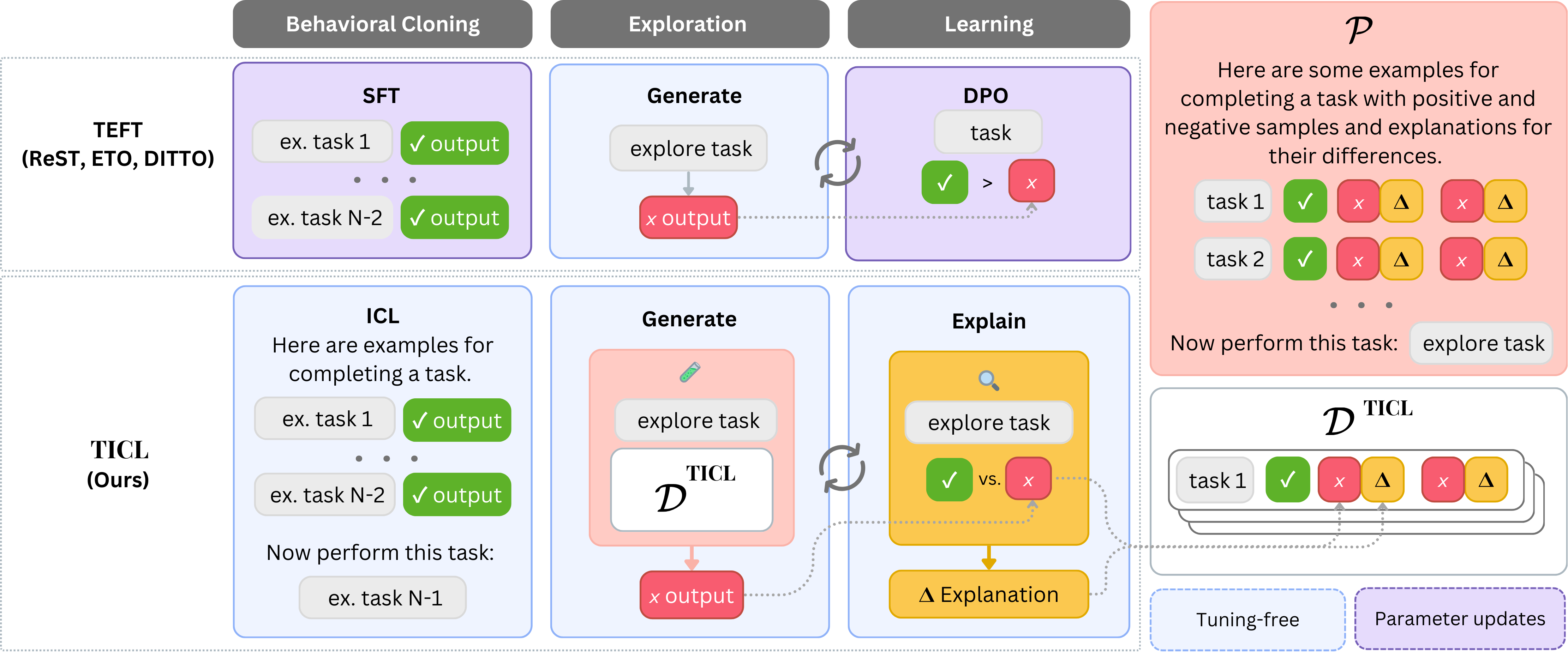}
    \caption{\ours methodology overview. Instead of SFT, \ours starts with a few-shot ICL for behavior cloning. 
    Then, \ours repeatedly generates an output for a task and a corresponding explanation that critiques the stylistic difference between the output and the user's text. If the output is considered not stylistically consistent, it and its explanation are added to the prompt.
    }
    \label{fig:method_overview}
\end{figure*}

\section{Methodology}
\label{sec:methodology}

Trial-Error-Explain In-Context Learning (\ours) is a method for expanding an ICL prompt through tuning-free correspondences of the three main stages in Trial-and-Error Fine-tuning (TEFT). 
Therefore, we describe \ours by going through each main stages of TEFT and the corresponding procedure in \ours, as illustrated in \autoref{fig:method_overview}.
TEFT refers to a method that (\textit{i}) fine-tunes a model with supervised fine-tuning (SFT) on a small set of training data (behavior cloning), (\textit{ii}) uses the trained model to generate outputs that are considered less preferred to those from the training data (exploration), (\textit{iii}) and fine-tunes the model further with a  preference optimization method, such as DPO (learning). 
This method has been applied to various tasks in previous work with minor differences, such as machine translation with ReST~\cite{gulcehre2023reinforced}, personalized text generation with DITTO~\cite{shaikh2024show}, and trajectory planning with ETO~\cite{song-etal-2024-trial}.

In \ours, we first replace SFT with ICL for behavior cloning (\textsection \ref{sec:behavioral_cloning}). 
Exploration is mostly the same except that \ours updates the prompt at the step level while TEFT updates the policy at the epoch level  in order to avoid augmenting the prompt with redundant errors (\textsection \ref{sec:exploration}). 
Lastly, for learning, we substitute preference optimization with prompt augmentation (\textsection \ref{sec:learning}). 
\ours generates explanations that analyzes the differences between the generated text and the user's actual text. Then, the generated text is labeled as a negative sample and the explanation are added to the prompt. 
We elaborate on each correspondence and share additional details in the following sections.

\subsection{Behavioral cloning}
\label{sec:behavioral_cloning}

As shown in the first step in \autoref{fig:method_overview}, TEFT uses supervised fine-tuning (SFT) to fine-tune a model $\pi_\theta$ roughly towards the desired behavior with the initial training data.
This helps with generating more relevant negative samples in the exploration phase~\cite{song-etal-2024-trial}. 

Formally, with the initial training data $\mathcal{D}^\mathcal{T}=\{(x_i,y_i)|_{i=1}^N\}$, where $x$ is the task and $y$ is the user's text, SFT uses the negative log likelihood (NLL) loss to learn from $\mathcal{D}^{\mathcal{T}}$ to fine-tune a pretrained LLM $\pi_\theta$ parameterized by $\theta$ such that: 
$$\mathcal{L}_{NLL}(\theta)=-\mathbb{E}_{(x,y)\sim\mathcal{D}^{\mathcal{T}}}\sum\limits_{t=1}^T\log\pi_{\theta}(y_t|y_{1:t-1},x)$$

\paragraph{SFT $\rightarrow$ Few-shot ICL.} 
To substitute SFT, we use $\mathcal{D}^{\mathcal{T}}$ to form an in-context learning (ICL) prompt and perform few-shot ICL ~\cite{tom2020fewshot}.
The resulting ICL prompt is defined as: 

\small
\begin{multline}
    \mathcal{P}\left(x_i,
    \mathcal{D}^{\texttt{icl}}\right),  \text{where} \\  \mathcal{D}^{\texttt{icl}} = \{(x_k,y_k,S_k)\}|_{k=1}^K, S = \{(\Tilde{y}_l, e_l)\}|_{l=1}^L
    \label{eq:our_prompt}
\end{multline}
\normalsize

Each sample in $\mathcal{D}^{\texttt{icl}}$ is a tuple of task $x$, user's text $y$, and a set of model-generated outputs and their corresponding explanations $S=\{(\Tilde{y}_{l}, e_{l})\}|_{l=1}^L$ from previous iterations $L$.  
$\mathcal{P}$ is a templating function that takes $\mathcal{D}^{\texttt{icl}}$, and the target input $x_i$ and forms an ICL prompt.\footnote{We share the details of $\mathcal{P}$ and all other prompts used in this work in Appendix \ref{appdx:prompt_details}.} 
In the first step of \ours, there are no previous model-generated outputs and explanations, so Eq.\ref{eq:our_prompt} is simply 
$\mathcal{P}\left(x_i,\{(x_k,y_k)|_{k=1}^K\}\right)$.

\subsection{Exploration}
\label{sec:exploration}

The objective of the exploration phase, the middle stage in \autoref{fig:method_overview}, is to generate negative samples that will be used for learning.
TEFT and \ours is mostly similar except for minor technical differences. 

In TEFT, the model  $\pi_\theta$ resulting from SFT or the previous iteration of DPO generates output $\Tilde{y}$ for the target task $x$ without any prompt templating. 
In contrast, \ours takes prompts resulting from Eq.\ref{eq:our_prompt} as input:
\begin{equation}
\Tilde{y_i}\sim \pi_\theta\left(p\left(x_i, \mathcal{D}^{\texttt{icl}}\right)\right) 
\label{eq:ours_explore}
\end{equation}

Another difference is that TEFT 
generates outputs for the full set of inputs in $\mathcal{D}^{\mathcal{T}}$: 
$\{\Tilde{y_i} \sim \pi_\theta(x_i)
| x_i \in \mathcal{D}^{\mathcal{T}}_0 \}$, prior to the learning stage. 
\ours only generates one output from a single sample before moving on to the learning phase -- think of $\text{batch size}=1$. 
We observe that generating outputs for all samples in $\mathcal{D}^{\mathcal{T}}$ leads to multiple $\Tilde{y}$ that contain redundant modes of failure. 
We empirically see that this inefficiency can be avoided by exploring a single $\Tilde{y}$ at a time.

\subsection{Learning}
\label{sec:learning}

In the learning phase for TEFT, the preference optimization samples $\mathcal{D}_i^{\texttt{pref}}$ collected from the exploration phase are used for fine-tuning $\pi_\theta$ with the use of a reference model $\pi_{\text{ref}}$, usually a pre-trained language model for preventing overfitting and reward hacking, and a preference optimization method, such as direct preference optimization (DPO)~\cite{rafailov2024direct}:

\paragraph{Preference optimization $\rightarrow$ Prompt augmentation with negative samples and explanations}

The main benefit of preference optimization is that the fine-tuned model learns a fine-grained understanding of the differences between reference outputs and negative samples. 
With \ours, we provide similar information in context via the negative samples and their corresponding explanations that analyze the difference between them and the reference outputs. 

We first let the model analyze the differences between the reference output $y$ and its respective negative output $\Tilde{y}$. 
This step is akin to rationale generation~
\cite{wang2022rationale, wei2022chain, zhou-etal-2022-reflect}, but the difference with prior work is that \ours's rationale focuses on explicitly extracting the differences between the reference output and the negative output, rather than producing a rationale as an intermediate step before fulfilling the target task. 

For this step, we generate explanations using an explanation generation prompt template $\mathcal{E}$ that takes in $y$ and $\Tilde{y}$:  
\begin{equation}
    (e_i,v_i)\sim f(\mathcal{E}\left(y_i, \Tilde{y_i})\right)
    \label{eq:explanation_generation}
\end{equation}

where $e_i$ is the explanation and $v_i$ is a validator Boolean that checks whether $\Tilde{y_i}$ is comparable in quality or accuracy as $y_i$. 
$f(\cdot)$ can be any function or combination of functions that generates $e_i$ and $v_i$. 
We use the same $\pi_\theta$ unless specified otherwise and have $\mathcal{E}$ instruct the model to generate both $e_i$ and $v_i$ in a JSON format. 
If $v_i\neq\text{True}$, then we add $(\Tilde{y_i},e_i)$ to $S_i$:  $(x_i,y_i)$'s set of model-generated outputs and explanations. 
Otherwise, we prevent good $\Tilde{y}$ from being included in the following iterations as negative samples, in order to avoid noise and the context becoming unnecessarily longer.

\paragraph{Iterations and checkpointing}

In summary, \ours starts with standard ICL to substitute SFT and  repeats (\textit{i}) exploration with an expanding dataset to  generate a candidate negative sample (Eq.\ref{eq:ours_explore}) and (\textit{ii}) learning to generate an explanation, expand the dataset with the negative sample and explanation if the sample is deemed a proper negative sample (Eq.\ref{eq:explanation_generation}).
A formal algorithm for \ours is defined in Algorithm \ref{alg:our_algorithm}. 
Denoting the augmented dataset that starts from $\mathcal{D}^{\mathcal{T}}$ as $\mathcal{D}^\texttt{\ours}$, 
we keep track of the version of $\mathcal{D}^\texttt{\ours}$ that attains the best performance on the validation set, measuring it every $n$ steps.

\RestyleAlgo{ruled}
\SetKwComment{Comment}{/* }{ */}

\begin{algorithm}[t!]
\small
\caption{
    \ours algorithm
}
\label{alg:our_algorithm}
\KwIn{$\pi_\theta$: LLM, $\mathcal{D}^\mathcal{T}=\{(x_i,y_i)|_{i=1}^N\}$: Initial training data, $L$: number of \ours iterations, $p(\cdot)$: ICL prompt formatting function, $S$: $\{(\Tilde{y}, e)\}$ where $\Tilde{y}$: negative sample and $e$: explanation.}
\KwOut{Augmented data $\mathcal{D}^{\text{\ours}}$}

$\mathcal{D}^{\text{\ours}} \gets \mathcal{D}^\mathcal{T}$

\For{$\text{l} \gets 0$ \KwTo $L$}{
    
    \For{$\text{i} \gets 0$ \KwTo $\left|\mathcal{D}^\mathcal{T}\right|-1$}{

        \textcolor{gray}{// Sample ICL examples excluding $x_i$ (\textsection \ref{sec:behavioral_cloning})} 
        
        $\mathcal{D^{\texttt{icl}}}=\texttt{random.sample(}\mathcal{D}^\mathcal{\text{\ours}}\setminus\{(x_i, y_i, S_i)\}\texttt{)}$
        
        \textcolor{gray}{ // Generate model output for $x_i$ (\textsection \ref{sec:exploration}})
        
        $\Tilde{y_i} \sim \pi_\theta\left(p(x_i, \mathcal{D^\texttt{icl}})\right)$

        \textcolor{gray}{// Generate explanation and determine whether it's a negative sample (\textsection \ref{sec:learning})}
        
        $(e_i, v_i) \sim \pi_\theta\left(\mathcal{E}(y_i, \Tilde{y}_i)\right)$ 

        \If{$v_i \neq \text{True}$}{
            \textcolor{gray}{// Update $\mathcal{D}^{\text{\ours}}$ with the negative sample and explanation}
            
            $S'_i \gets S_i \cup \{(\Tilde{y_i}, e_i)\} $  

            $(x_i, y_i, S_i) \in \mathcal{D}^{\text{\ours}}, (x_i, y_i, S_i) \mapsto (x_i, y_i, S'_i)$
        
        }
    }
}

\Return $\mathcal{D}^{\text{\ours}}$
\end{algorithm}

\section{Experimental Setup}

\subsection{Evaluation Dataset}
\label{sec:dataset}
We build on the evaluation setup from DITTO~\cite{shaikh2024show}, which uses two public authorship attribution (AA) datasets, because they also study personalized alignment. 
AA is a classification task of predicting the correct author for some written document, but it can also be reformulated as a personalized alignment task of generating text that is attributed to a specific author.  
Same as DITTO, we consider 20 authors from two sources:
(\textit{i}) CMCC~\cite{goldstein2008cmcc}, which contains emails and essays on controversial topics written by students, and (\textit{ii}) CCAT~\cite{lewis2004ccat50}, which consists of news articles written by journalists.
We preprocess the data with the same steps as DITTO to simulate a realistic data scenario for personalization in which per-user data is limited and treated as private.  
The number of samples per author and per task is small~\cite{zhang2020improved}, limited to the smallest number of samples available for an author (12) across both datasets to control for data size. 
In addition, each author's data is used in isolation such that performance does not depend on the total amount of data available among all authors. 
The train/val/test split is set to 7/2/3.\footnote{Our processed data can be found in \url{https://justin-cho.com/ticl}.}

\subsection{Evaluation Method}
\label{sec:pairwise_evaluation}

\paragraph{Pairwise comparison on stylistic similarity with LLM-as-a-judge}

\begin{figure}[t]
    \centering
    \includegraphics[width=0.9\linewidth]{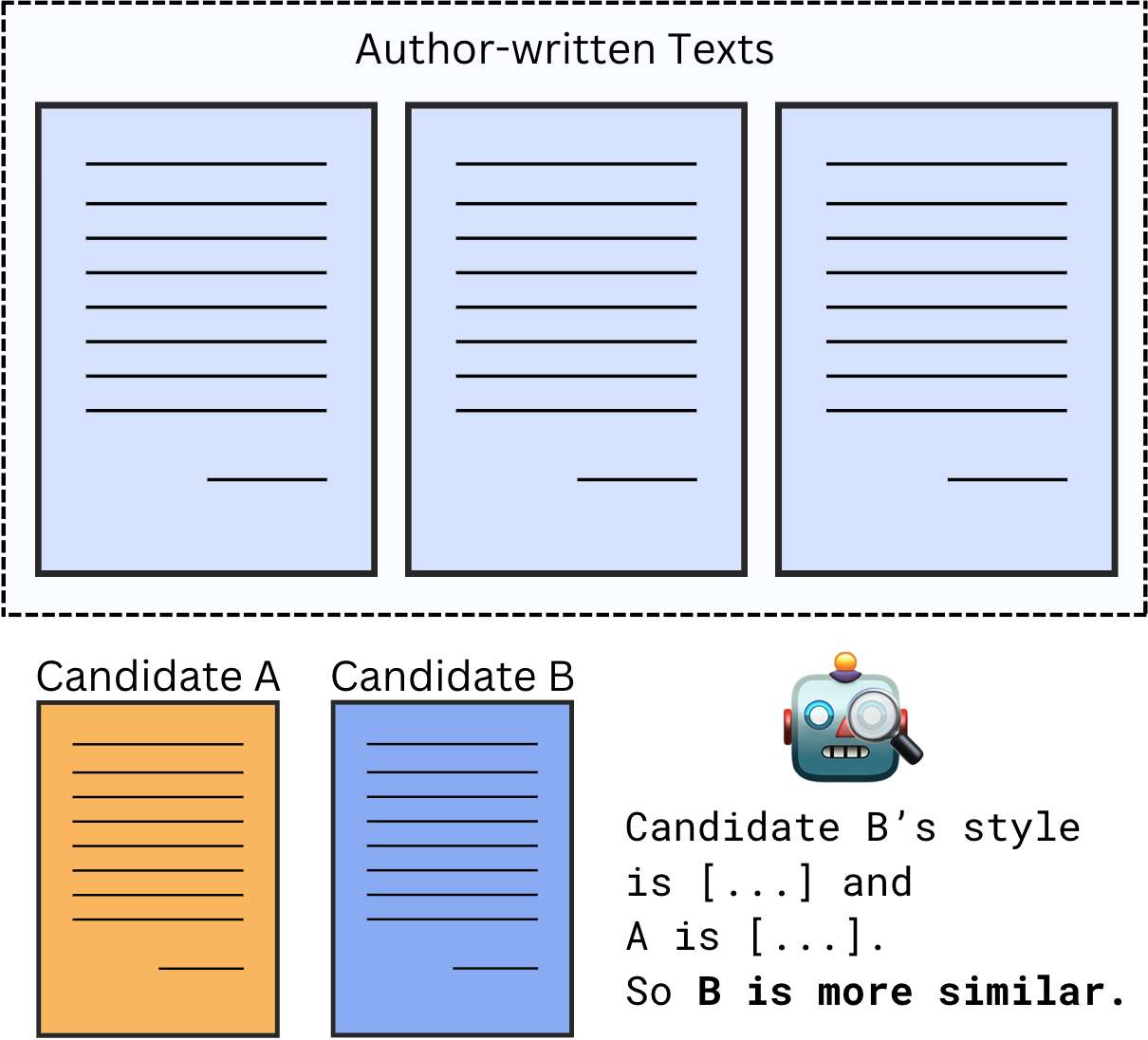}

    \caption{Simplified illustration of our evaluation setup for pairwise comparison of stylistic similarity using LLM-as-a-judge. The full prompt is show in \autoref{fig:eval_template}.}
    \label{fig:simple_evaluation_prompt}
\end{figure}

We evaluate model performance with pairwise comparisons of generated text using LLM-as-a-judge~\cite{zheng2023judging, kim2023prometheus}, asking which of the given candidates is more similar to a set of user's texts. 
A simplified illustration of our evaluation approach is shown in \autoref{fig:simple_evaluation_prompt}.

Specifically, we ask GPT-4o\footnote{The version we use is \texttt{gpt-4o-2024-0806}.} which of two candidates outputs generated for an author's prompt in the test set is more stylistically consistent with five examples sampled from the author's training data.
This is similar to the method used in DITTO~\cite{shaikh2024show}, but we found that we can further increase accuracy by providing more examples, providing a list of style elements to focus on, and generating an explanation before making its final decision.  
The two main comparisons we are interested in are how our methods and baselines compare with DITTO's outputs and the author's text. 
The former is a comparison against the prior state-of-the-art and the latter is against the gold output, and therefore serves as an upper bound. 
In other words, achieving a $\geq50\%$ win rate against the author indicates that the method is able to produce outputs that are indistinguishable from the author's texts for our LLM-as-a-judge in terms of stylistic consistency.

\begin{table}[t]
    \centering
    \resizebox{\linewidth}{!}{
    \begin{tabular}{lrr}
        \toprule 
         Benchmark& Config & Acc.   \\ 
         \midrule
         \multirow{2}{*}{CMCC}  & all & $89.5_{2.3}$ \\ 
          & Top 10 & $96.6_{1.0}$ \\ 
          \midrule 
          \multirow{4}{*}{CCAT} & all & $90.1_{2.6}$  \\  
          & Top 10 & $98.1_{0.7}$ \\ 
          & all, TF-IDF-controlled &  $80.5_{2.0}$ \\  
          & Top 10, TF-IDF-controlled & $96.8_{1.2}$ \\ 
          \bottomrule
         
    \end{tabular}
    }
    \caption{Accuracy for our LLM-as-a-judge setup on author texts. We use the top 10 authors who are most reliably identified by our LLM-as-a-judge to get more reliable evaluation results. Results for individual authors are shown in in appendix in \autoref{tab:per_author_evaluation_benchmark}.}
    \label{tab:llm_as_a_judge_main}
\end{table}

\paragraph{Evaluating LLM-as-a-judge for  pairwise comparisons of stylistic similarity\footnote{We experiment with various automatic evaluation metrics, human evaluation and alternative LLM-as-a-judge setups, and find pairwise comparisons to be the most reliable. Refer to Appendix \ref{appdx:automatic_evaluation} for these results. 
We intentionally avoid human evaluation based on our own results of human performance on comparing style similarity and evidence from previous work have shown that model-based classifiers are more reliable at determining stylistic similarity than untrained humans~\cite{hallinan-etal-2023-steer, krishna-etal-2020-reformulating, liu2024authorshipstyletransferpolicy, liu2024styletransfermultiiterationpreference}}.}

We validate the accuracy of our evaluation method by benchmarking it with user data in CMCC and CCAT. 
For CMCC, one of the two candidate outputs is the actual text from the correct author and the other is from another human author who wrote a response for the same prompt. 
This way we can prevent the LLM-as-a-judge from using topical similarity for determining correct authorship, since both candidates will discuss similar content \cite{wegmann-etal-2022-author}. 
For CCAT, there are no texts written for the same prompt, so we find the most similar distractor among texts from all other authors using TF-IDF. 
With this setup, our pairwise comparison evaluation with LLM-as-a-judge achieves $\sim97\%$ accuracy for both CMCC and CCAT for the top 10 authors, as shown in \autoref{tab:llm_as_a_judge_main}.

\paragraph{Author selection} 
Based on these benchmarking results, 
we discover significantly lower accuracies for some authors, as shown by the large drop in accuracy in \autoref{tab:llm_as_a_judge_main} when all authors are considered. 
We hypothesize that these authors have less consistent style across their texts, which may cause inconsistent evaluation results. Therefore, instead of random sampling, we select the top 10 authors in terms of our LLM-as-a-judge's accuracy from each dataset for a total of 20 authors.

\begin{table*}[t]
    \centering
    \small
\begin{tabular}{@{}lll|cc|cc@{}}
\toprule
& & & \multicolumn{2}{c|}{Win rate vs. DITTO} & \multicolumn{2}{c@{}}{Win rate vs. Author} \\
 \cmidrule(lr){4-5} \cmidrule(l){6-7}
\multicolumn{3}{c|}{\textbf{Method}} & \textbf{CMCC} & \textbf{CCAT} & \textbf{CMCC} & \textbf{CCAT} \\
\midrule
Mistral & \multicolumn{2}{l|}{DITTO} & - & - & $25.00_{1.08}$ & $10.00_{0.55}$ \\
\midrule
\multirow{5}{*}{GPT-4o} & \multicolumn{2}{l|}{Zero-shot} & $9.00_{0.72}$ & $21.00_{1.05}$ & $2.50_{0.34}$ & $6.00_{0.40}$ \\
& \multicolumn{2}{l|}{Few-shot} & $42.50_{1.13}$ & $76.50_{1.25}$ & $21.00_{1.28}$ & $25.50_{0.56}$ \\
& \multicolumn{2}{l|}{CoT} & $41.00_{1.01}$ & $74.50_{1.04}$ & $12.00_{0.53}$ & $21.50_{0.47}$ \\
& \multicolumn{2}{l|}{OPRO} & $3.00_{0.56}$ & $13.50_{1.27}$ & $0.00_{0.00}$ & $0.00_{0.00}$ \\
& \multicolumn{2}{l|}{\textbf{\ours}} & $\mathbf{53.00_{1.24}}^{\dag}$ & $\mathbf{82.50_{1.25}}^{\dag}$ & $\mathbf{31.00_{1.18}}^{\dag}$ & $\mathbf{28.00_{0.63}}^{\dag}$ \\
\midrule
\multirow{5}{*}{Claude 3 Sonnet} & \multicolumn{2}{l|}{Zero-shot} & $12.00_{1.03}$ & $54.00_{1.44}$ & $0.50_{0.11}$ & $10.50_{0.66}$ \\
& \multicolumn{2}{l|}{Few-shot} & $74.00_{1.51}$ & $86.50_{0.69}$ & $43.50_{1.37}$ & $53.50_{0.94}$ \\
& \multicolumn{2}{l|}{CoT} & $71.50_{1.20}$ & $90.00_{0.67}$ & $44.00_{0.88}$ & $51.00_{0.81}$ \\
& \multicolumn{2}{l|}{OPRO} & $2.00_{0.25}$ & $23.50_{1.33}$ & $0.00_{0.00}$ & $0.00_{0.00}$ \\
& \multicolumn{2}{l|}{\textbf{\ours}} & $\mathbf{79.50_{1.26}}^{\dag}$ & $\mathbf{91.50_{0.74}}$ & $\mathbf{54.50_{1.67}}^{\dag}$ & $\mathbf{57.00_{0.87}}^{\dag}$ \\
\bottomrule
\end{tabular}

    \caption{Win rates and their corresponding standard errors as subscripts against DITTO~\cite{shaikh2024show} and the original author's response. $\dag$ indicates a win rate that is larger than the next best performing baseline at a $p<0.05$ statistically significant level.
    \ours outperforms all baselines at a statistically significant level, except for Claude 3 Sonnet on CCAT when compared to CoT.
    Win rate is averaged across 10 authors from each dataset. Full results with per-author scores are in \autoref{tab:main_results_vs_ditto} and \autoref{tab:main_results_vs_author}.}
    \label{tab:main_combined}
\end{table*}

\subsection{Baselines}
\label{sec:baselines}

To measure the relative performance of \ours, we also evaluate on 
(\textit{i})\textbf{ Zero-shot},  to capture the base model's vanilla behavior when it is only given the task as input, and 
(\textit{ii}) \textbf{ICL}, where $N$ samples of task and output pairs of the author are included in the context as examples to learn from, along with some guidance to focus on style elements.\footnote{The full set of prompts we use in our work can be found in Appendix \ref{appdx:prompt_details}.} 

In addition, we evaluate other methods that scale test-time compute to improve downstream performance. 
(\textit{iii}) \textbf{Chain-of-Thought}~\cite{wei2022chain} is a prompting method that asks to reason about the style elements in the few-shot examples and then generate the response according to the analyzed style and the few-shot examples.  
(\textit{iv}) \textbf{OPRO}~\cite{yang2024large}is a prompt optimization method that refines the prompt iteratively using chain-of-thought prompt as the initial prompt. The model self-updates its prompt based on the history of previous prompts and their scores as ``gradients'' that guide the update. 
The initial prompt is set to ``Let's think step by step''~\cite{kojima2022large} and the scores are cosine similarity between the Style Embeddings (SE)~\cite{wegmann-etal-2022-author} of the generated outputs and the reference texts of the validation set. 
The prompt with the highest score is used at test time.

We also compare \ours to
(\textit{v}) \textbf{DITTO}~\cite{shaikh2024show}, the state-of-the-art that uses TEFT on a small model, i.e. Mistral 7B Instruct~\cite{jiang2023mistral}, and outperforms GPT-4 combined with few-shot ICL. 
We do not include SFT in our experiments because DITTO was already shown to be superior to SFT.
Lastly, we also compare to (\textit{vi}) \textbf{Author}, the author's actual response to the given prompts and thus the upper bound.
For (\textit{i})-(\textit{iv}), we use \texttt{gpt-4o-2024-08-06}~\cite{achiam2023gpt} and \texttt{Claude 3 Sonnet}~\cite{TheC3}, abbreviated henceforth as GPT-4o and Claude 3 S unless otherwise specified. 
We let DITTO and \ours to run for four epochs. For \ours, we use LLM-as-a-judge between outputs from the validation set to choose the best performing prompt. 
We use Mistral 7B Instruct v0.2 for DITTO, following the original implementation.

\section{Results and Discussion}
\label{sec:results}
\subsection{Main Results}
\label{ssec:main_results}
The main evaluation results that compare the baselines against DITTO and against the author's actual reference text is shown in \autoref{tab:main_combined}.
Details on the sample size can be found in Appendix \ref{appdx:sampling}.

\paragraph{Vs. DITTO} \ours outperforms or performs on par with all other baselines and often outperforms DITTO by a significant margin. 
We find that explicitly analyzing writing style in the few-shot examples and asking the model to abide by these styles did not lead to consistent improvements. 
With the exception of Claude 3 Sonnet on CCAT, CoT actually underperforms the simpler few-shot setup.
Interestingly, we find that OPRO consistently performs worst. 
When we analyzed the optimization sequence, OPRO lands on prompts that look sensible, e.g. \textit{``Respond with rhetorical devices, simple sentence structures, and easy vocabulary...}, but these generalized guidance are not as effective as seeing few-shot examples within the context in producing stylized text.  

\paragraph{Vs. Author} We see similar strengths for \ours when compared to Author, for which the results are shown in \autoref{tab:main_results_vs_author}. 
Surprisingly, Claude 3 Sonnet is able to exceed the $\geq 50$ win rate threshold when using \ours for both CMCC and CCAT. 
Performance with GPT-4o is relatively lower, but still significantly higher than DITTO for both datasets. 
We further dive into why Claude 3 Sonnet outperforms GPT-4o in \textsection \ref{sec:gpt_vs_claude}

Given Claude's strong performance even in the zero-shot setting against DITTO, we suspect there may be some leakage of CCAT data into Claude.  
Therefore, we focus our analysis henceforth mainly on CMCC. 
Also, since the relative performance among models is similar between vs. DITTO and vs. Author results, we focus on vs. Author results in the following sections.

\begin{table}[t]
    \small
    \centering
    \resizebox{\linewidth}{!}{
    \begin{tabular}{llr}
         \toprule
         \textbf{Model} & \textbf{Ablation} & \multicolumn{1}{c}{\textbf{Win rate}}  \\
         \midrule 
         \multirow{5}{*}{GPT-4o} & \ours & $\mathbf{31.00_{1.18}}$ \\
         & $-$ Initial ICL examples & $28.50_{1.50}$\\ 
         & $-$ Explanations & $23.50_{1.24}$ \\ 
         & $-$ Checkpointing & $22.50_{0.90}$ \\ 
         & $-$ Negative samples \& Expl. & $21.00_{1.28}$ \\ 
         \midrule 
         \multirow{5}{*}{Claude 3 S} & \ours & $\mathbf{54.50_{1.67}}$ \\ 
          & $-$ Initial ICL examples & $52.00_{1.22}$\\ 
         & $-$ Explanations & $46.00_{1.41}$ \\ 
         & $-$ Checkpointing & $54.00_{1.67}$ \\ 
         & $-$ Negative samples \& Expl. & $43.50_{1.37}$ \\ 
         \bottomrule
    \end{tabular}
    }
    \caption{Ablation with \ours minus ($-$) individual procedures on CMCC. Reported win rate is vs. Author and subscript is the standard error. All procedures provide performance gains, but explanations make the biggest difference while starting with ICL examples is the least important. 
    }
    \label{tab:ablation_results_cmcc}
\end{table}

\begin{table}[t]
    \small
    \centering
    \resizebox{\linewidth}{!}{
    \begin{tabular}{llr}
         \toprule
         \textbf{Model} & \textbf{Swap Configuration} &  \multicolumn{1}{c}{ Win rate}   \\
         \midrule 
         \multirow{3}{*}{GPT-4o} & No swap & $31.00_{1.18}$ \\ 
         & $\rightarrow$ with Claude 3 S expl. & $32.50_{1.13}$ \\ 
         & $\rightarrow$ with Claude 3 S \ours   & $18.30_{0.25}$ \\
         \midrule
         \multirow{3}{*}{Claude 3 S} & No swap    & $54.50_{1.67}$ \\ 
         & $\rightarrow$ with GPT-4o expl. & $55.00_{1.28}$ \\ 
         & $\rightarrow$ with GPT-4o \ours  & $42.50_{0.92}$ \\ 
         \bottomrule
    \end{tabular}
    }
    \caption{Ablation with various swap configurations to compare GPT-4o \ours and Claude 3 Sonnet \ours performance on CMCC.
    Result format is the same as \autoref{tab:ablation_results_cmcc}.
    Changing the explanation model does not lead to significant changes. 
    Using final \ours prompts from a different model leads to large drops, indicating that models learn better from their own failure modes.
    }
    \label{tab:gpt_vs_claude}
\end{table}

\begin{table*}[]
\small
    \centering
    \adjustbox{max width=\textwidth}{    \begin{tabular}{llrrlrr}
    \toprule
         \multicolumn{1}{c}{\textbf{A}} & \multicolumn{1}{c}{\textbf{B}} & \textbf{FRE ($\text{A}\setminus \text{B}$)} $\uparrow$ & \textbf{FRE ($\text{B}\setminus \text{A}$)} $\uparrow$ & \textbf{$n$-gram} & \textbf{$z$-score} & $p$ \\ \midrule 
         \multirow{4}{*}{GPT-4o \ours} & \multirow{4}{*}{GPT-4o Few-shot} & \multirow{4}{*}{121.22} & \multirow{4}{*}{36.62} & so why & $1.990$ & 0.047 \\  
         & & & & honestly & $2.098$ & 0.036 \\ 
         & & & & additionally & $-2.983$ & 0.003 \\
         & & & & therefore & $-2.233 $ & 0.026 \\ 
         \midrule 
         \multirow{4}{*}{Claude 3 S \ours} & \multirow{4}{*}{GPT-4o \ours} & \multirow{4}{*}{$120.21$} & \multirow{4}{*}{$77.21$} & believe & 4.255 & 0.000 \\
         & & & & feel that & 4.195 & 0.000  \\
         & & & & crucial to & $-3.449$ & 0.001 \\ 
         & & & & implications of & $-2.925$ & 0.003 \\ 
         \bottomrule
    \end{tabular}
    }
    \caption{Representative Fightin' Words model~\cite{monroe2008fightin} results for frequency differences between two sources at $p<0.05$. A positive z-score indicates that output source A significantly uses more of the n-gram than B while a negative score signals the opposite. 
    We observe that the models that have higher win-rates from pairwise comparisons frequently generate phrases for expressing personal opinions while the losing model generates more formal phrases. 
    }
    \label{tab:fightin_words_model_results}
\end{table*}

\subsection{Ablating \ours}

We conduct an ablation study to understand how each of the components in \ours contributes to its performance. 
We observe how the win rates change when we remove the initial ICL setup that substitutes SFT for behavior cloning, explanations, checkpointing (choosing the best performing augmented prompt throughout iterations rather than simply using the final prompt), and both the negative samples and explanations.
The last row is equivalent to the few-shot baseline. 
The results are shown in \autoref{tab:ablation_results_cmcc}. 

\paragraph{Explanations are most important, while starting with ICL provides smallest improvements.}
For both GPT-4o and Claude 3, removing any of the procedures leads to performance drops. 
For both cases, we see the largest drop attributed to removing both the explanations and negative samples. 
When only dropping the explanations, we see a smaller drop, indicating that both the negative samples and explanations provide additive advantages. 
For GPT-4o, checkpointing is also very important, but less so for Claude 3 Sonnet. 
In both models, not having the first step use ICL examples is not critical, meaning that we can even start with zero-shot outputs for the first step and not see large performance degradation.

\subsection{Why does Claude 3 Sonnet outperform GPT-4o?}
\label{sec:gpt_vs_claude}

One striking trend from \autoref{tab:main_combined} is Claude 3 Sonnet's noticeably stronger performance compared to GPT-4o. 
This is somewhat surprising given that GPT-4o is the latest model that ranks at the top of popular language modeling leaderboards, such as Chatbot Arena.\footnote{\url{https://lmarena.ai/}}
In order to understand Claude's comparative performance over GPT-4o, we examine how performance changes if we (\textit{i}) use the other model for generating explanations during prompt augmentation (with other model expl.) and (\textit{ii}) use the fully developed \ours prompt from the other model at test time (with other model \ours).  

\paragraph{Explanations are similar in quality.}
Results shown in \autoref{tab:gpt_vs_claude} suggest that it is \textbf{not} that Claude 3 Sonnet generates better explanations than GPT-4o. 
Applying \ours on GPT-4o with Claude 3 Sonnet as the explanation generator leads to a mild increase in performance that is not statistically significant. 
This is further corroborated by Claude 3 Sonnet's results when using GPT-4o as the explanation generator, which also leads to a statistically insignificant improvement. 

\paragraph{\ours prompts are not transferable.}
Using the final augmented prompts developed by Claude 3 Sonnet and using GPT-4o at test-time does not improve on simply using GPT-4o's own augmented prompt. 
This is likely due to the distribution shift: negative samples and their corresponding feedback is not as relevant and actually serves as noise such that the performance actually drops below the few-shot performance in \autoref{tab:main_combined}. 
Based on these results, we attribute Claude 3 Sonnet's superior performance on its superior few-shot ICL capabilities for writing tasks. 
The following section on lexical analysis adds support to this hypothesis. 

\subsection{Lexical Analysis}
\label{sec:lexical_analysis}

We further explore the models' comparative performances using a lexical analysis, using 
Fightin' Words model for extracting words with significant differences in between two corpora~\cite{monroe2008fightin}. The method shows different ratios of words that are significantly more frequent in one model over another. 
Using a $p$-level of 0.05, we calculate the number of $n$-grams that are significantly more frequent in the author's text compared to the model's text and vice-versa.
Representative results are shown in \autoref{tab:fightin_words_model_results}.

\paragraph{\ours overcomes model bias for structural and formal phrases better than ICL.}
Comparing GPT-4o \ours vs. GPT-4o few-shot elicits \ours improvement over a simple ICL baseline. 
We see that structural and formal phrases, e.g. ``\textit{additionally}'' and ``\textit{therefore}'', are more frequent in the few-shot outputs, while \ours contain more colloquial phrases, e.g., ``\textit{so why}''. 
This suggests that \ours enables models to overcome its biases towards generic formal text and better learn from the given style context. 

\paragraph{Claude 3 Sonnet applies user styles more effectively than GPT-4o.}
When comparing Claude 3 Sonnet and GPT-4o directly, the top words that Claude 3 Sonnet is able to generate more consistently are subjective and assertive words or phrases that are frequently seen in CMCC, e.g., ``\textit{believe}'' and ``\textit{feel that}'', and also more casual phrases, e.g., ``\textit{things like}'' and ``\textit{kind of}.''  
On the other hand, GPT-4o's outputs consists of more formal phrases, such as ``\textit{crucial to}.'' 
This dynamic is further corroborated by the Flesch Readability Ease (FRE) scores ~\cite{flesch1948new}.
The words that appear significantly more frequently for Claude has a higher FRE score (FRE(A$\setminus$B)$=$120.21) than the score for words that appear significantly more frequenty for GPT-4o (FRE(B$\setminus$A$=$77.21). 
A higher FRE score means it is easier to understand when read, which correlates to more casual text~\cite{cho2024speechworthyinstructiontunedlanguagemodels}.

\section{Related Work}

\subsection{Personalization}

Personalization is an overloaded term in the context of language models as it can refer to multiple distinct features. 
Language models can be personalized to complete tasks as a specific user~\cite{salemi2023lamp, zhuang2024hydramodelfactorizationframework}, answer questions according to a user's profile~\cite{kim2024fewshotpersonalizationllmsmisaligned, dutt-etal-2022-perkgqa}, and recommend content based on a user's knowledge level~\cite{wang2016personalized}. They can also respond in a way that aligns with an individual's preference of digital assistance behavior~\cite{jang2023personalized}, edit an existing text to match someone else's style~\cite{horvitz-etal-2024-tinystyler, patel2023lowresourceauthorshipstyletransfer} . and generate some output in the style of a target user, such as a response to another online user~\cite{liu-etal-2023-recap} or text for a writing task~\cite{shaikh2024show, kumar2024longlampbenchmarkpersonalizedlongform, li2024learning, mysore2023pearl}. 

\ours focuses on the last category mentioned above -- personalized text generation -- where we want to steer language models away from a generic output~\cite{li-etal-2016-diversity, zhang-etal-2021-trading, padmakumar2024does} for a given writing task and instead produce one that is specific to a user's style~\cite{rivera-soto-etal-2021-learning, wegmann-etal-2022-author}. 
Prior work on personalized text generation rely on at least one component that requires fine-tuning, either the model that generates the response~\cite{li2023teach, liu-etal-2023-recap, shaikh2024show, mysore2023pearl} or a model that revises the prompt passed on to a black box model~\cite{li2024learning}. 
However, \ours is an inference-only method that does not require any parameter updates or new parameters. 

\subsection{Inference-only learning}

One major branch of inference-only learning method focuses on automating prompt engineering, optimizing the wording of a prompt template that is placed prior to asking a model to complete a task~\cite{shin-etal-2020-autoprompt, zhou2023large, ma2024largelanguagemodelsgood, yang2024large, kim2024fewshotpersonalizationllmsmisaligned, ye-etal-2024-prompt}, e.g. \textit{``Let's think step by step''} from \cite{kojima2022large}. 
Recent work have operationalized this process to also automate example selection for few-shot examples~\cite{yuksekgonul2024textgrad, khattab2023dspy}. 
In contrast, \ours maximizes its understanding of the target task by iteratively augmenting the content used for ICL with contrastive examples and explanations while keeping the prompt template constant.

Another prominent branch of inference-only learning is agentic methods that build on intermediate LLM outputs to refine the final output~\cite{chen-etal-2023-self, wei2022chain, yao2023react, sumers2024cognitive, shinn2024reflexion, saha-etal-2024-branch}. 
Reflexion~\cite{shinn2024reflexion} is most similar to our work, but it cannot be directly applied without a reliable evaluator for measuring success. 
Another key difference with \ours is that it front-loads the trial-and-error process ahead of test time and does not augment the prompt a priori by reasoning over few-shot examples.

\section{Conclusion}

In conclusion, we develop \ours, a tuning-free method for personalized alignment that adapts prior work on trial-and-error fine-tuning with scaling inference compute.   
Instead of fine-tuning a model with synthetic negative samples, \ours uses them to augment the prompt and generate explanations that provide further fine-grained guidance on how to align outputs towards a desirable style.
On personalized text generation tasks, \ours outperforms other competitive tuning-free baselines and the previous state-of-the-art.  
\ours provides an approach for leveraging black box models for personalization without any fine-tuning.

\section*{Limitations} 

One of the limitations of \ours is that it is more computationally costly than fine-tuned models at test time because of the longer inputs due to the few-shot examples and their corresponding negative samples and explanations. 
However, inference cost for repeating prompts can be significantly lowered by prompt caching.\footnote{\url{https://www.anthropic.com/news/prompt-caching}} 
In addition, inference costs in general are continuously getting lower with optimization efforts that comprehensively span hardware\footnote{\href{https://groq.com/}{Groq} and  \href{https://aws.amazon.com/machine-learning/inferentia/}{AWS Inferentia} are some examples of inference-specific hardware development.} and software~\cite{kwon2023efficient}. 
As the promise of inference-only methods become increasingly evident, we believe these efforts will be further accelerated and lead to even more favorable conditions for methods such as \ours. 

Another limitation of \ours is that it is only effective for models that are able to understand long contexts well. 
In our preliminary experiments, our results on using \ours with smaller models such as Mistral 7B Instruct and GPT-4o mini were poor as these models continued to generate outputs similar in style to what were labeled in-context as bad examples, indicating that they were not able to distinguish between the good and bad examples included in the context.
Therefore, the performance of \ours is dependent on a model's long context understanding. 

Lastly, \ours results in per-user or per-task prompts, which is similar to the limitations of prior work that rely on fine-tuned per-user model parameters. 
Whether we can use \ours to enable a single model to perform personalized task completion for multiple tasks without requiring excessively long prompts by (\textit{i}) discovering methods that maintain or improve performance while compressing the prompt or (\textit{ii}) injecting the information in the prompt for controlling style during decoding is an exciting avenue for future work.

\section*{Generative AI Statement}
No generative AI tools were used in the writing of this work.

\bibliography{anthology,custom}
\bibliographystyle{acl_natbib}

\appendix

\section*{Appendix}
\label{sec:appendix}

\section{Evaluation Dataset Details}
\label{appdx:dataset_details}

\subsection{Sampling}
\label{appdx:sampling}
We rely on sampling to alleviate issues of having a small test set per author. 
For each of the three test prompts, we sample five outputs per method. 
We pair outputs for the same prompts to create 75 comparison pairs ($5\cdot5\cdot3$) and sample 40 of them for comparison and randomly assign the order they appear in the evaluation prompt to control for order bias. 
For the versus author results, we only have 15 samples since there’s only one original author response ($1\cdot5\cdot3)$, so we permute the order for each pair to get 30 samples. 
This results in 400 ($40\cdot10$) comparisons for vs DITTO and 300 ($30\cdot10$) for vs Author to get the averaged win rates in \autoref{tab:main_combined}. 
We are able to compute statistically significant win rates with these number of comparisons, as reported in Section \ref{ssec:main_results}.

\subsection{Technical Details}
\label{appdx:technical_details}

The exact models that we use for our experiments are \texttt{gpt-4o-2024-0806}, Claude 3 Sonnet and Mistral 7B Instruct v0.2 via Amazon Bedrock. 
For evaluation (\textsection \ref{sec:pairwise_evaluation}), we use batch inference using the OpenAI API, which lets us reduce inference costs by 50\%.\footnote{\url{https://platform.openai.com/docs/guides/batch}} 

\paragraph{Notes on \ours}
At the first step of \ours, we hold out one sample $(x_i, y_i)$ at random and use the rest   $\mathcal{D}^{\mathcal{T}}_0\setminus\{(x_i, y_i)\}$ as few-shot examples that form the ICL prompt.
If we use the full set, we don't have any samples for which we can explore and learn  without seeing the reference output $y_i$ of the input $x_i$. from $\mathcal{D}^{\mathcal{T}}_0$.
This information is made more explicit in Algorithm \ref{alg:our_algorithm}.

\paragraph{Notes on DITTO.} 
DITTO was trained with Mistral 7B Instruct, which is suspected to be a smaller model than Claude 3 Sonnet and GPT-4o. 
We found that the fine-tuned variants of Mistral 7B Instruct are prone to generating unstable outputs, producing template components such as \texttt{[INST]} or generating degenerate outputs that contain repetitive content. 
To give these approaches the best chances possible against our approaches with larger models, we reject samples that contain template components and repetitive content until we reach the desired number of samples $N=5$ from each model.

\section{Evaluation Methods}
\label{appdx:automatic_evaluation}

\begin{table}
    \centering
    \resizebox{\linewidth}{!}{
    \begin{tabular}{llrrrrrrrrrr}
    \toprule
        Benchmark & \multicolumn{10}{c}{Author IDs}  \\ 
        \midrule
        CMCC &   2 & 3 & 5 & 6 & 7 & 8 & 11 & 13 & 15 & 17 \\ 
        CCAT & 10 & 12 & 15 & 20 & 23 & 27 & 28 & 30 & 32 & 38    \\ 
        \bottomrule 
    \end{tabular}
    }

    \caption{Mapping from author indices to actual author IDs in CMCC and CCAT that we use throughout our work. These authors are the authors that ranks in the top 10 for highest accuracy achieved with GPT-4 Eval, as described in Appendix \ref{appdx:llm as a judge} and shown in \autoref{tab:per_author_evaluation_benchmark}.}
    \label{tab:author_ids}    
\end{table}

\begin{table}[t]
    \centering
    \resizebox{\linewidth}{!}{
    \begin{tabular}{lrrrr}
        \toprule 
         Method & \# Examples & Explan. & CMCC & CCAT  \\ 
        \midrule
        \multirow{4}{*}{GPT-4o pairwise} & 1 & \xmark & $76.1_{2.2}$ & -  \\ 
        & 1 & \cmark & $78.1_{2.1}$ & - \\ 
        & 3 & \cmark & $82.6_{1.9}$ & - \\ 
        & 5 & \cmark & $89.5_{2.3}$ &  ${80.4}_{2.0}$\\ \midrule
        GPT-4o mini pairwise & 5 & \cmark & $68.2_{2.4}$ & $66.8_{2.3}$ \\ \midrule 
        GPT-4o rating & 5 & \cmark & $55.0_{1.2}$ & $20.0_{1.1}$ \\ \midrule 
        \multirow{3}{*}{SE} & 1 & - & $71.5_{1.5}$ & $61.3_{1.5}$ \\
         & 3 & - & $76.8_{1.4}$ & $63.0_{1.5}$ \\
        & 5 & - & $78.5_{1.3}$ &  $63.6_{1.5}$ \\ \midrule
        Human & 5 & - & $85.4$ & - \\ 
        \bottomrule
    \end{tabular}
    }
    \caption{Results with author texts from CMCC and CCAT for comparing various evaluation methods. GPT-4o: \texttt{gpt-4o-2024-0806}, GPT-4o mini: \texttt{gpt-4o-2024-0718-mini}, SE: Style Embeddings.}
    \label{tab:eval_method_comparison}
\end{table}

In this section, we describe benchmarking results using author texts from CMCC and CCAT for various evaluation methods. 
Results are shown in \autoref{tab:eval_method_comparison}.
For CMCC, several authors wrote responses to the same prompt, and therefore we use samples written for the same prompt as the compared samples, as this would help us control for content similarity between the tested sample and the examples in making the task arbitrarily easy. 

For CCAT, which are news articles with a format of ``\textit{Write a news article that starts with the following sentence}: \texttt{article's first sentence}''. 
There are no share prompts for this dataset. 
Since journalists tend to write articles on similar topics over time and thus content similarity can be an easy hint for detecting authorship, we select the negative sample as one that has the highest TF-IDF similarity with any one of the in-context examples.

\subsection{Embedding-based evaluation}

Universal authorship representations (UAR)~\citep{rivera-soto-etal-2021-learning} is a sentence embedding method that uses a SBERT model~\cite{reimers-gurevych-2019-sentence} fine-tuned with a contrastive learning objective such that the representations of sentences or documents from the same author become closer together than with those of other authors.
Style Embeddings (SE) is a follow-up to UAR that refines the representations to focus on style rather than content by pairing contrastive samples that share similar content~\citep{wegmann-etal-2022-author}, and therefore we focus on embedding-based comparisons with Style Embeddings. 

We compute the cosine similarity between candidate texts' SE with those of the author examples and the one with higher cosine similarity is considered more stylistically similar. 
The best score for SE is $78.5\%$ and $63.6\%$ for CMCC and CCAT, respectively, which is significantly lower than GPT-4o results but better than GPT-4o mini's results.
While having more examples also helps with SE's performance, it plateaus markedly at three examples, and there is only minimal gains observed for CCAT.

\subsection{LLM-as-a-Judge}
\label{appdx:llm as a judge}

The pairwise comparison setup is described in Section \ref{sec:pairwise_evaluation}. 
As an alternative, we also consider having LLM-as-a-judge provide ratings for individual instances, so that we can reduce the number of samples that the LLM-as-a-judge need to evaluate and also measure relative performance of each apporach based on their aggregate scores. 
The prompt is similar to the pairwise comparison setup except that only one candidate is shown and the LLM is asked to provide a score from 1-5 on how stylistically similar it thinks the candidate is to the examples. 
Unfortunately, this result is the poorest performing approach, only achieving 55.0\% accuracy for CMCC and 20.0\% accuracy for CCAT. 
The main reason for low accuracy is that the majority of instances were given the same rating such that they were deemed as ties.

\subsection{Human Evaluation}
\label{appdx:human_evaluation}
We have three human annotators complete the same setup with 100 samples for CMCC only with five samples from each author. 
Conducting the same evaluation for CCAT was overwhelming for human participants because the average text length in CCAT is much larger.
As shown in the results, human performance is lower than the best GPT-4o results. 
This reinforces findings from previous work that identifying authorship reliably is a difficult task for untrained humans and that model-based classifiers are more reliable and practical for this task~\cite{hallinan-etal-2023-steer, krishna-etal-2020-reformulating, liu2024authorshipstyletransferpolicy, liu2024styletransfermultiiterationpreference}.

The full prompt template for this evaluation setup is shown in \autoref{fig:eval_template}.

\section{Flesch Reading Ease}
\label{appdx:flesch}

The Flesch Reading Ease (FRE) score is a simple equation for approximating readability of a given text based on the average number of words per sentence and the average number of syllables per word~\cite{flesch1948new}.
The formula for calculating FRE is the following: 

$$
FRE= 206.835 - (1.015\cdot ASL)-(84.6\cdot \frac{n_{sy}}{n_w})
$$

where $ASL$ is average sentence length ($\frac{\text{total words}}{\text{total sentences}}$) and $\frac{n_{sy}}{n_w}$ is the average number of syllables per word ($\frac{\text{total syllables}}{\text{total words}}$).  
The higher the score, the easier it is to understand a piece of text when read. 
The maximum score is 121.22, while there is no limit to how low it can be and it can even be negative. 
We use Python's \texttt{textstat} package\footnote{\url{https://github.com/textstat/textstat}} to compute FRE.

\begin{table*}[t]
    \centering
    \small
    \begin{tabular}{l|llr|rrrrrrrrrr}
    \toprule
         Data & \multicolumn{2}{c}{Method} & \multicolumn{1}{c}{$a_{avg}$} & $a_1$ & $a_2$ & $a_3$ & $a_4$ & $a_5$ & $a_6$ & $a_7$ & $a_8$ & $a_9$ & $a_{10}$ \\   
         \midrule 
         \multirow{10}{*}{CMCC} & \multirow{5}{*}{GPT-4o} & Zero-shot & $9.00_{0.72}$ & 20 & 5& 0 & 0 & 15 & 10 & 10 & 0 & 30& 0 \\
         & &  Few-shot & $42.50_{1.13}$ & 55& 35& 40& 20& 65& 55& 40& 15& 55& 45 \\ 
         & & CoT  & $41.00_{1.01}$ & 50& 35& 35& 25& 55& 50& 50& 15& 60& 35\\ 
         & & OPRO & $3.00_{0.56}$ & 5& 0& 0& 0& 0& 0& 0& 0& 25& 0\\ 
         & & \ours & $\mathbf{53.00_{1.24}}^{\dag}$ & 70& 70& 50& 35& 60& 65& 80& 45& 70& 50 \\  \cmidrule{2-14}
         & \multirow{5}{*}{Claude 3 Sonnet} & Zero-shot & $12.00_{1.03}$ & 25& 5& 5& 0& 20& 15& 5& 0& 45& 0 \\
         & &  Few-shot & $74.00_{1.51}$ & 85& 55& 70& 35& 85& 90& 95& 50& 100& 75 \\ 
         & & CoT & $71.50_{1.20}$ & 80& 60& 65& 55& 70& 85& 90& 45& 100& 65\\ 
         & & OPRO & $2.00_{0.25}$ & 5& 0& 0& 0& 0& 0& 5& 0& 10& 0\\ 
         & & \ours & $\mathbf{79.50_{1.26}}^{\dag}$ & 85& 50& 65& 90& 75& 95& 90& 95& 100& 100 \\ 
         \midrule 
         \multirow{10}{*}{CCAT} &  \multirow{5}{*}{GPT-4o} & Zero-shot & $21.00_{1.05}$ & 15& 20& 15& 5& 50& 5& 30& 20& 40& 10 \\
         & &  Few-shot & $76.50_{1.25}$ & 90& 70& 90& 40& 90& 70& 95& 90& 60& 70 \\ 
         & & CoT & $74.50_{1.04}$ & 75& 65& 85& 45& 85& 90& 90& 90& 80& 40 \\ 
         & & OPRO & $13.50_{1.27}$ & 0& 5& 0& 0& 35& 10& 55& 15& 10& 5\\ 
         & & \ours & $\mathbf{82.50_{1.25}}^{\dag}$ & 85& 65& 90& 70& 85& 90& 100& 90& 80& 70 \\ \cmidrule{2-14}
         & \multirow{5}{*}{Claude 3 Sonnet} & Zero-shot & $54.00_{1.44}$ & 50& 40& 65& 35& 65& 35& 90& 80& 50& 30\\
         & &  Few-shot&  $86.50_{0.69}$ & 95& 80& 95& 65& 95& 85& 95& 85& 90& 80\\ 
         & & CoT & $90.00_{0.67}$ & 100& 95& 100& 70& 85& 85& 100& 90& 90& 85 \\ 
         & & OPRO &$23.50_{1.33}$ & 20& 5& 10& 10& 55& 5& 55& 15& 30& 30  \\ 
         & & \ours & $\mathbf{91.50_{0.74}}$ & 100& 90& 100& 75& 100& 90& 100& 95& 80& 85\\  
         \bottomrule
         
    \end{tabular}
    \caption{LMJ-based win rate results against DITTO~\cite{shaikh2024show}. $\dag$ indicates a win rate that is larger than the next best performing baseline at a $p<0.05$ statistically significant level.
    \ours outperforms all other baselines at a statistically significant level, except for Claude 3 Sonnet on CCAT. $a_{avg}$ is the average win rate across all authors and $a1$-$a10$ are the author IDs from \autoref{tab:author_ids}.}
    \label{tab:main_results_vs_ditto}
\end{table*}

\begin{table*}[t]
    \centering
    \small
    \begin{tabular}{l|llr|rrrrrrrrrr}
    \toprule
         Data & \multicolumn{2}{c}{Method} & \multicolumn{1}{c}{$a_{avg}$} & $a_1$ & $a_2$ & $a_3$ & $a_4$ & $a_5$ & $a_6$ & $a_7$ & $a_8$ & $a_9$ & $a_{10}$ \\   
         \midrule 
         \multirow{10}{*}{CMCC} & Mistral & DITTO & $25.00_{1.08}$ & 20& 20& 35& 45& 10& 35& 5& 30& 5& 45\\
         \cmidrule{2-14}
         & \multirow{5}{*}{GPT-4o} & Zero-shot & $2.50_{0.34}$ & 0& 0& 15& 5& 0& 0& 0& 0& 5& 0 \\
         & &  Few-shot & $21.00_{1.28}$ & 20& 15& 10& 25& 35& 15& 10& 0& 15& 65\\ 
         & & CoT  & $12.00_{0.53}$ & 25& 10& 10& 10& 20& 5& 10& 0& 10& 20\\ 
         & & OPRO & $0.00_{0.00}$ & 0& 0& 0& 0& 0& 0& 0& 0& 0& 0\\ 
         & & \ours & $\mathbf{31.00_{1.18}}^{\dag}$ & 30& 20& 15& 35& 35& 25& 45& 5& 35& 65\\  \cmidrule{2-14}
         & \multirow{5}{*}{Claude 3 Sonnet} & Zero-shot & $0.50_{0.11}$ & 0& 0& 0& 0& 0& 0& 0& 0& 5& 0\\
         & &  Few-shot & $43.50_{1.37}$ & 60& 40& 30& 25& 45& 45& 55& 5& 60& 70 \\ 
         & & CoT & $44.00_{0.88}$ & 50& 40& 30& 45& 25& 40& 65& 35& 55& 55 \\ 
         & & OPRO & $0.00_{0.00}$ & 0& 0& 0& 0& 0& 0& 0& 0& 0& 0\\ 
         & & \ours & $\mathbf{54.50_{1.67}}^{\dag}$ & 80& 15& 30& 75& 60& 75& 65& 55& 60& 85\\ 
         \midrule 
         \multirow{11}{*}{CCAT} & Mistral & DITTO & $10.00_{0.55}$ & 0& 15& 15& 15& 5& 25& 0& 5& 10& 10 \\
         \cmidrule{2-14}   
         &  \multirow{5}{*}{GPT-4o} & Zero-shot & $6.00_{0.40}$ & 0& 10& 0& 15& 10& 0& 0& 10& 10& 5 \\
         & &  Few-shot & $25.50_{0.56}$ & 15& 15& 30& 25& 35& 35& 15& 30& 25& 30\\ 
         & & CoT & $21.50_{0.47}$ & 15& 10& 15& 20& 25& 30& 25& 30& 20& 25 \\ 
         & & OPRO & $0.00_{0.00}$ & 0& 0& 0& 0& 0& 0& 0& 0& 0& 0 \\ 
         & & \ours & $\mathbf{28.00_{0.63}}^{\dag}$ & 10& 30& 25& 40& 45& 40& 20& 30& 30& 30 \\ \cmidrule{2-14}
         & \multirow{5}{*}{Claude 3 Sonnet} & Zero-shot & $10.50_{0.66}$ & 0& 5& 10& 25& 20& 0& 0& 20& 15& 10 \\
         & &  Few-shot&  $53.50_{0.94}$ & 60& 70& 40& 55& 45& 40& 35& 55& 60& 75\\ 
         & & CoT & $51.00_{0.81}$ & 55& 70& 40& 50& 50& 35& 45& 50& 45& 70  \\ 
         & & OPRO &$0.00_{0.00}$ & 0& 0& 0& 0& 0& 0& 0& 0& 0& 0  \\ 
         & & \ours & $\mathbf{57.00_{0.87}}^{\dag}$ & 60& 55& 45& 55& 80& 40& 45& 55& 70& 65 \\  
         \bottomrule
         
    \end{tabular}
    \caption{LLM-as-a-judge win rates against the author's actual text. The table takes the same format as \autoref{tab:main_results_vs_ditto}.}
    \label{tab:main_results_vs_author}
\end{table*}

\section{Full Per-Author Results}

The full per-author results with comparisons against DITTO and against the author's text are shown in \autoref{tab:main_results_vs_ditto} and \autoref{tab:main_results_vs_author}, respectively.

\section{Sample Outputs}

We show some qualitative examples from our baselines and \ours in \autoref{tab:sample_outputs}. 

\begin{table*}[h]
    \footnotesize
    \centering
    \adjustbox{max width=\textwidth}{
        \begin{tabular}{p{0.5\textwidth}| p{0.5\textwidth}}
            \toprule
            \multicolumn{1}{c}{\textbf{CMCC}} & \multicolumn{1}{c}{\textbf{CCAT}} \\
            \midrule
            \multicolumn{2}{c}{Dataset} \\ 
            \midrule
            Write an approximately 500 word essay to the following prompt: Recently, school officials prevented a school shooting because one of the shooters posted a myspace bulletin. Do you think this was an invasion of privacy? & Write an article that starts with the following: A potential new source of revenue -- oil -- offers hope of a fresh start for Chad's fragile economy as it rebuilds after three decades of war. \\ 
            \midrule
            \multicolumn{2}{c}{Author}\\
            \midrule
            \char91...\char93~I feel as though the school officials were not invading privacy at all. 
            The entire point
            ~\char91...\char93~
            Why anyone~\char91...\char93~
            is beyond me, and even more so if the person does not expect to be caught. If anyone~\char91...\char93~
            Did the school harm the person who was going to do the shooting? No. ~\char91...\char93
            since it IS available to the public- to TURN OVER ~\char91...\char93~
            If there are lives in danger~\char91...\char93~
            (to catch kids drinking, etc.), then THAT would be "taking it over the edge" as an invasion of privacy. Thanks for your time, ID \# 20
            & 
            \char91...\char93~``Soon work will begin,'' Foreign Minister Saleh Kebzabo told Reuters recently. ``Chad should get some royalties so we can free the country through development.'' 
            Chad has known coups, civil war or conflict with its northern neighbour Libya for much of its life from independence in 1960. Sporadic cross-border rebel raids remain a problem.
            \char91...\char93~
            But
            \char91...\char93~
            On the negative side, external debt rose to \$817 million in 1996 from \$803 million in 1995 and 1996 GDP growth was projected at 0.5 percent, down from 4.5 percent in 1995. 
            ~\char91...\char93~
            \\
            \midrule
            \multicolumn{2}{c}{DITTO}\\ \midrule 
            \char91...\char93~I definitely do not think that the school system invaded the privacy of the person who was caught posting the myspace bulletin about the potential school shooting. 
            ~\char91...\char93~
            If this was the case, then he definitely knows what a good portion of the world is aware of- the fact that everything you do can be caught on camera. Whether it be a store, etc., everything is being recorded. 
            ~\char91...\char93~
            Sincely, ID\# 20123452
& 
\char91...\char93~
The landlocked African nation tapped its first well in 1973 but suffered instability through French suitor Cameroon's push for a larger share of any wealth or former guerrilla revolts against president Idriss Deby's rule. ``Oil is a godsend. It is not only a source of revenue but it also gives Chad prestige on the international scene,'' said Jean-Pierre Alain, head of Chad's national hydrocarbons agency. 
~\char91...\char93~
It had an estimated 1997/98 budget of \$275 million. It had international debt of \$355 million at the end of 1996
~\char91...\char93~

\\
            \midrule
            \multicolumn{2}{c}{Few-shot}
            \\ \midrule 
            \char91...\char93~ In regards to the question of whether the school officials' actions to prevent a school shooting based on a myspace bulletin were an invasion of privacy, I hold the firm belief that they were not. The fundamental duty of any educational institution ~\char91...\char93~
            Ultimately, ~\char91...\char93~
            Hence, ~\char91...\char93~ Thank you for your inquiry. -ID\#: 20
& 
            \char91...\char93~The discovery of significant oil reserves in the Doba Basin has invigorated economic forecasts and raised hopes of alleviating poverty in the landlocked Central African nation. 
            \"This is an opportunity for Chad to transform its economy,\" said Finance Minister Alhabo Abakar. \"The potential revenues from oil could be a game-changer.\"
            ~\char91...\char93~
            The oil pipeline, a multi-billion-dollar joint venture with global oil firms, promises to open up isolated regions and create jobs. 
            \char91...\char93~
            \\ \midrule
            \multicolumn{2}{c}{CoT}\\ \midrule 
            \char91...\char93~I firmly believe that the actions taken by school officials in this situation were not an invasion of privacy.
            In fact, 
            ~\char91...\char93~
            In light of this, ~\char91...\char93~
            As www.dictionary.com defines a "bulletin,"
            ~\char91...\char93~
            Hence,~\char91...\char93~
            Thank you for reaching out to discuss this crucial topic. ID\#: 20
 & 
\char91...\char93~``Oil could be the lifeline we've been waiting for,'' said Economic Minister Mahamat Djoum. 
 ``It has the potential to transform our economy, create jobs, and improve the standard of living for our people.''
 Chad's oil reserves, located primarily in the Doba basin in the south, are expected to yield significant output, with projections suggesting several hundred thousand barrels a day by the year 2000. 
 \char91...\char93~
            \\ \midrule
            \multicolumn{2}{c}{OPRO} \\ \midrule  
            Thank you for reaching out with this important question. While privacy is a significant concern,
            ~\char91...\char93~
            Nonetheless, it's crucial to continue the discussion on maintaining a balance between security and individual rights.
            & 
            \char91...\char93~Historically plagued by persistent conflict and political instability, Chad has struggled to maintain sustainable economic growth and improve the quality of life for its citizens. 
            \char91...\char93~
            \\   
            \midrule
            \multicolumn{2}{c}{\ours}\\ \midrule 
            \char91...\char93~ I genuinely believe that ~\char91...\char93~ It's simple - ~\char91...\char93~ If someone posts about harmful intentions, expecting privacy is a bit ironic, don't you think? The school officials did their job, ~\char91...\char93~ Privacy is essential, yes, ~\char91...\char93~ If a message was genuinely private, it wouldn’t be on such a public platform, plain and simple. So, let’s focus ~\char91...\char93~ Remember, if you put it out there, it’s open to be acted upon for the greater good. Thanks for hearing me out on this matter! 
            &
            \char91...\char93~``Oil represents a new chapter for Chad's economy, promising the possibility of financial stability and development,'' said Finance Minister Mahamat Allamine Bourma.
            The nation, which remains one of the world's poorest despite its natural resources, eyes the oil sector with optimism. 
            The World Bank has been involved in supporting infrastructure projects, including the Chad-Cameroon pipeline, designed to facilitate the transport of oil to international markets.
            \char91...\char93~ 
            A consortium led by ExxonMobil, Chevron, and Malaysia's Petronas plans to invest over \$3 billion            ~\char91...\char93~
               \\
            \bottomrule
        \end{tabular}
    }
    \caption{Truncated sample responses from baselines and \ours using GPT-4o for CMCC and CCAT. 
    }
    \label{tab:sample_outputs}
\end{table*}

\section{Prompt Details}
\label{appdx:prompt_details}

For details on the prompts that we use, we refer the reader to the following: 

\begin{itemize}
    \item Few-shot prompting: \autoref{fig:fewshot_prompt}
    \item \ours: \autoref{fig:our_prompt}
    \item Pairwise LMJ evaluation: \autoref{fig:eval_template}
    \item Style guide generation for CoT: \autoref{fig:cot_style_guide_template}
    \item Writing prompt for CoT that uses the generated style guide: \autoref{fig:cot_writing_template}
    \item OPRO's optimization prompt: \autoref{fig:opro_optimization_template}
    \item  OPRO's writing prompt that uses the prompt found from the optimization process: \autoref{fig:opro_writing_template}
\end{itemize}

\begin{figure*}
    \centering
\begin{tcolorbox}[width=\textwidth]
\fontsize{8pt}{8pt}\selectfont\ttfamily 
You are an editor. \\ 
\begin{itemize}
\renewcommand\labelitemi{--}
    \item Your task is to analyze whether the candidate writing is stylistically consistent with the author's writing(s) and if not, highlight elements of the author's style that are not observed in the candidate writing. 
    \item Consider similarity with regards to the (1) length, (2) format, (3) paragraph structure, (4) sentence structure, (5) punctuation, (6) syntax, (7) voice, and (8) diction of the author's writing, but NOT the content it covers. 
    \item Use the minimum words possible in your analysis while providing specific examples of how the observed inconsistencies must be edited to become stylistically consistent with the author's writing.
    \item If the candidate writing is stylistically consistent with the author's writing, respond with "yes" in the "is\_consistent" field. Otherwise, respond with "no". 
\end{itemize}
\vspace{2mm}

\# Task \\ 
\{ task \} \\ 
\vspace{2mm}

\# Author's writing  \\ 
\{ reference\_text \} \\ 
\vspace{2mm}

\# Candidate writing to edit \\ 
\{ generated\_text \} \\ 
\vspace{2mm}

Respond only with JSON with the following format:\\ 
\{ \\ 
\phantom{xx}"explanation": "<style analysis and suggested edits>",\\ 
\phantom{xx}"is\_consistent": "<yes/no>"\\ 
\} 

\end{tcolorbox}

    \caption{Prompt template $\mathcal{E}$ for generating explanations on the difference between the reference output and the generated output.}
    \label{fig:explanation_prompt}
\end{figure*}

\begin{figure*}
    \centering
\begin{tcolorbox}[width=\textwidth]
\fontsize{8pt}{8pt}\selectfont\ttfamily
You are a stylistically consistent writer. Below are examples that exemplify your writing style.

\vspace{2mm}

\# Writing Task Example 1 \\ 
\{ example["task"] \} \\ 
\#\# Your Writing 1 \\ 
\{ example["reference\_output"] \} \\ 

\# Writing Task Example 2 \\ 
... \\ 
\vspace{2mm}

** Task to complete ** \\ 
Now complete the following writing task with a style and format consistent with `Your Writing` examples.
\\
Be consistent in terms of (1) length, (2) format, (3) paragraph structure, (4) sentence structure, (5) punctuation, (6) syntax, (7) voice, and (8) diction of your writing when completing the task. 
\\
\vspace{2mm}

Task: \{ target\_task \} \\ 

Directly provide your response in the following format:\\ 
\begin{verbatim} 
```
<your writing>
```
\end{verbatim}

\end{tcolorbox}
    \caption{Prompt template $\mathcal{P}$ for the few-shot in-context learning baseline.}
    \label{fig:fewshot_prompt}
\end{figure*}

\begin{figure*}
    \centering
\begin{tcolorbox}[
 width=1.0\textwidth
]
\fontsize{8pt}{8pt}\selectfont
\ttfamily
You are a stylistically consistent writer. Below are examples that exemplify your writing style.

\vspace{2mm}

\# Writing Task Example 1 \\ 
\{ example["task"] \} \\ 
\#\# Your Writing 1 \\ 
\{ example["reference\_output"] \} \\ 

\#\# Stylistically Inconsistent Writing 1-1 \\ 
\textcolor{blue}{\{ example["generated\_output"][0]["output"] \}} \\ 
\#\#\# Inconsistent stylistic elements in `Stylistically Inconsistent Writing 1-1` that should be corrected for it to become more consistent with `Your Writing 1`: \\ 
\textcolor{darkgreen}{\{ example["generated\_output"][0]["explanation"] \}} \\ 
... \\

\# Writing Task Example 2 \\ 
... \\ 
\vspace{2mm}

** Task to complete ** \\ 
Now complete the following writing task with a style and format consistent with `Your Writing` examples and also avoiding the stylistic inconsistencies found in the `Stylistically Inconsistent Writing` examples.
\\
Be consistent in terms of (1) length, (2) format, (3) paragraph structure, (4) sentence structure, (5) punctuation, (6) syntax, (7) voice, and (8) diction of your writing when completing the task. 
\\
\vspace{2mm}

Task: \{ target\_task \} \\ 

Directly provide your response in the following format: 
\begin{verbatim} 
```
<your writing>
```
\end{verbatim}

\end{tcolorbox}
    \caption{Prompt template $\mathcal{P}$ for \ours. It builds on \autoref{fig:fewshot_prompt} by adding \textcolor{blue}{negative samples} and \textcolor{darkgreen}{explanations} generated with \autoref{fig:explanation_prompt}.}
    \label{fig:our_prompt}
\end{figure*}

\begin{figure*}
    \centering
\begin{tcolorbox}[
width=1.0\textwidth
]
\fontsize{8pt}{8pt}\selectfont
\ttfamily
You are an impartial evaluator of style similarity. Below are samples of an author's writing and two options. 

\vspace{2mm}

\# Author's Writing:\\ 
EXAMPLE 1: \\ 
\{example 1\}\\ 
...\\ 
EXAMPLE 5: \\ 
\{example 5\} 

\vspace{2mm}

\# Option A:

\{candidate A\}

\vspace{2mm}

\# Option B:

\{candidate B\}

\vspace{2mm}

\# Task

Which option is more likely to have been written by the author based on style similarity to the samples given as AUTHOR'S WRITING above? Consider each option's similarity with regards to the (1) length, (2) format, (3) paragraph structure, (4) sentence structure, (5) punctuation, (6) syntax, (7) voice, and (8) diction of the author's writing, but NOT the content it covers. If one option has incoherent/odd text or formatting (e.g., random dashes, repetitive text, random signatures, etc.) that is not present in the author's writing while the other doesn't, it should be considered less similar.

\{ \\
\phantom{xx}"explanation": \{ \\ 
\phantom{xx}\phantom{xx}"length": "<Explanation>", \\ 
\phantom{xx}\phantom{xx}    "format": "<Explanation>", \\ 
\phantom{xx}\phantom{xx}    "paragraph structure": "<Explanation>", \\ 
\phantom{xx}\phantom{xx}    "sentence structure": "<Explanation>", \\ 
\phantom{xx}\phantom{xx}    "punctuation": "<Explanation>", \\ 
\phantom{xx}\phantom{xx}    "syntax": "<Explanation>", \\ 
\phantom{xx}\phantom{xx}    "voice": "<Explanation>", \\ 
\phantom{xx}\phantom{xx}    "diction": "<Explanation>", \\ 
\phantom{xx}\phantom{xx}    "odd incoherent text/formatting": "<Explanation>" \\ 
\phantom{xx} \} \\ 
\phantom{xx}"answer":  "<The option more similar to the AUTHOR'S WRITING; either A or B>" \\ 
\}

\vspace{2mm}

ALWAYS REMAIN IMPARTIAL WHEN EVALUATING OUTPUTS AND PENALIZE ODD INCOHERENT TEXT.
\end{tcolorbox}
    \caption{Prompt template for evaluating which output is more stylistically similar to the target author's writing style given five examples from the target author.}
    \label{fig:eval_template}
\end{figure*}

\begin{figure*}
    \centering
\begin{tcolorbox}[width=\textwidth]
\fontsize{8pt}{8pt}\selectfont\ttfamily 
Given the following written examples, provide a style guide on how you would go about writing in a similar style as the examples for the target task. Analyze the (1) length, (2) format, (3) paragraph structure, (4) sentence structure, (5) punctuation, (6) syntax, (7) voice, and (8) diction of the examples, but NOT the content or topic that they cover. \\ 

\vspace{2mm}

\# Writing Task Example 1 \\ 
\{ example["task"] \} \\ 
\#\# Your Writing 1 \\ 
\{ example["reference\_output"] \} \\ 

\# Writing Task Example 2 \\ 
... \\ 
\vspace{2mm}

\# Target task \\ 
Task: \{ target\_task \} 

\vspace{2mm}

Provide your style guide with the following format:
\begin{verbatim}
```
<your writing>
```
\end{verbatim}

\end{tcolorbox}

    \caption{Prompt template for generating style guides that will be used for Chain-of-Thought guided writing in \autoref{fig:cot_writing_template}.}
    \label{fig:cot_style_guide_template}
\end{figure*}

\begin{figure*}
    \centering
\begin{tcolorbox}[width=\textwidth]
\fontsize{8pt}{8pt}\selectfont\ttfamily 
You are a stylistically consistent writer. Below are examples that exemplify your writing style. \\ 
\vspace{2mm}

\# Writing Task Example 1 \\ 
\{ example["task"] \} \\ 
\#\# Your Writing 1 \\ 
\{ example["reference\_output"] \} \\ 

\# Writing Task Example 2 \\ 
... \\ 
\vspace{2mm}

\# Task to complete \\ 
Now complete the following writing task, first by analyzing the style and format of the `Your Writing` examples. \\ 

Task: \{ target\_task \} \\ 

\textcolor{blue}{\{ Chain-of-Thought Style Guide \}} \\ 

Directly provide your response for the task in the following format:
\begin{verbatim}
```
<your writing>
```
\end{verbatim}

\end{tcolorbox}

    \caption{Prompt template for Chain-of-Thought-guided writing. The \textcolor{blue}{blue text} is the style guide generated from \autoref{fig:cot_style_guide_template}.}
    \label{fig:cot_writing_template}
\end{figure*}

\begin{figure*}
    \centering
\begin{tcolorbox}[width=\textwidth]
\fontsize{8pt}{8pt}\selectfont\ttfamily 
I have some instructions along with their corresponding scores. The instructions are arranged in ascending order based on their scores, where higher scores indicate better quality. \\ 
\vspace{2mm}

\begin{verbatim}
{%
instruction: 
{{ prompt_score_pair[0] }}
score:
{{ prompt_score_pair[1] }}
{%
\end{verbatim}

The following exemplars show how to apply your instruction: you replace <INS> in each input with your instruction, then read the input and give an output. The scores are the average stylistic similarity score of the generated output, generated with your instruction, compared to the reference output.

\begin{verbatim}
{%
input:
{{ example["task"] }}
<INS>
output:
{ example["reference_output"] }
{%
\end{verbatim}

\vspace{2mm}

Write your new instruction that is different from the old ones that will help achieve a style similarity score as high as possible. Consider similarity with regards to the (1) length, (2) format, (3) paragraph structure, (4) sentence structure, (5) punctuation, (6) syntax, (7) voice, and (8) diction of exemplar outputs, but NOT the content it covers. The instruction should not mention 'reference output' or 'original input' as only the input and the instruction placed instead of <INS> is available when writing an output. Write your instruction in the following format:
\begin{verbatim}
```
<your instruction>
```
\end{verbatim}

\end{tcolorbox}

    \caption{OPRO optimization prompt. The instruction with the highest score is used for the OPRO writing prompt in \autoref{fig:opro_writing_template}.}
    \label{fig:opro_optimization_template}
\end{figure*}

\begin{figure*}
    \centering
\begin{tcolorbox}[width=\textwidth]
\fontsize{8pt}{8pt}\selectfont\ttfamily 
Task: \{ target\_task \} \\ 
\textcolor{blue}{\{ opro\_prompt \}}

\begin{verbatim}
Respond only with JSON with the following format:
{
    "thought": "<your thoughts>",
    "response": "<your response>"
}
\end{verbatim}

\end{tcolorbox}

    \caption{OPRO writing prompt. The \textcolor{blue}{blue text} is the instruction that is selected from the prompt optimization process from \autoref{fig:opro_optimization_template}.}
    \label{fig:opro_writing_template}
\end{figure*}

\begin{table*}[t]
    \centering
    \begin{tabular}{lrrr}
        \hline
        Author ID & CMCC & \makecell{CCAT\\w/o TF-IDF} & \makecell{CCAT\\w/ TF-IDF} \\
        \hline
        0 & 84 & 96 & 82 \\
        1 & 74 & 96 & 76 \\
        2 & 98 & 92 & 74 \\
        3 & 92 & 98 & 74 \\
        4 & 72 & 92 & 88 \\
        5 & 94 & 64 & 58 \\
        6 & 98 & 100 & 80 \\
        7 & 96 & 100 & 86 \\
        8 & 92 & 68 & 70 \\
        9 & 88 & 68 & 64 \\
        10 & 90 & 100 & 100 \\
        11 & 98 & 86 & 86 \\
        12 & 84 & 100 & 98 \\
        13 & 100 & 96 & 76 \\
        14 & 90 & 96 & 68 \\
        15 & 100 & 100 & 96 \\
        16 & 64 & 84 & 82 \\
        17 & 98 & 96 & 80 \\
        18 & 88 & 92 & 68 \\
        19 & - & 78 & 76 \\
        20 & - & - & 100 \\
        21 & - & - & 58 \\
        22 & - & - & 70 \\
        23 & - & - & 92 \\
        24 & - & - & 64 \\
        25 & - & - & 60 \\
        26 & - & - & 90 \\
        27 & - & - & 98 \\
        28 & - & - & 100 \\
        29 & - & - & 90 \\
        30 & - & - & 90 \\
        31 & - & - & 80 \\
        32 & - & - & 100 \\
        33 & - & - & 68 \\
        34 & - & - & 76 \\
        35 & - & - & 86 \\
        36 & - & - & 72 \\
        37 & - & - & 60 \\
        38 & - & - & 94 \\
        39 & - & - & 88 \\
        \hline
        Accuracy & $89.47_{2.0}$ & $90.05_{1.7}$ & $80.45_{1.7}$ \\
        \hline
    \end{tabular}
    \caption{Full benchmarking results using human written texts in CMCC and CCAT for LLM-as-a-judge for style similarity, using \texttt{gpt-4o-2024-0806}. Values shown are per-author accuracies (\%). We used only the first 20 authors for comparing CCAT with and without using TF-IDF. 
    After finding that TF-IDF helps control for semantic similarity among author examples, we expanded the evaluation using TF-IDF to all authors to find authors that our evaluation achieves the highest accuracy.}
    \label{tab:per_author_evaluation_benchmark}
\end{table*}

\end{document}